\pdfoutput=1
\documentclass[11pt,letterpaper]{article}
\usepackage[margin=1in]{geometry}

\usepackage{times}
\usepackage{epsfig}
\usepackage{graphicx}
\usepackage{amsmath}
\usepackage{amssymb}
\usepackage{booktabs}
\usepackage{textcomp}
\usepackage{subcaption}
\usepackage{soul}
\usepackage{algorithm}
\usepackage{algpseudocode}
\usepackage{paralist}

\usepackage[inline]{enumitem}
\usepackage[table,dvipsnames]{xcolor}
\usepackage{multirow}
\usepackage{tabularx}
\usepackage{tablefootnote}
\usepackage{url}
\usepackage[colorlinks]{hyperref}

\newcommand{\eg}{\textit{e.g.}, }

\newcommand{\starnote}[1]{}
\newcommand{\laststarnote}[1]{}
\newcommand{\donestarnote}[1]{}

\makeatletter
\DeclareRobustCommand{\change}{\@bsphack\normalcolor\@esphack}
\DeclareRobustCommand{\jchange}{\@bsphack\normalcolor\@esphack}
\DeclareRobustCommand{\stopchange}{\@bsphack\normalcolor\@esphack}
\makeatother

\newcommand{\blfootnote}[1]{%
  \begingroup\renewcommand\thefootnote{}\footnotetext{#1}\addtocounter{footnote}{-1}\endgroup}

\title{\bf Leveraging Cost-effective Robotics for K-12 STEM Education\\through Water Quality Monitoring Tasks}

\author{
Rishi Mukherjee$^{1}$, Andrew Ruiz$^{2}$, Travis Henderson$^{3}$, Resha Tejpaul$^{4}$, Kris Simonson$^{5}$,\\
David Mulla$^{6}$, Brian McNeil$^{7}$, Nikolaos Papanikolopoulos$^{8}$, Junaed Sattar$^{9}$\\[1ex]
{\small $^{1,3,4,6,7,8,9}$Dept. CSE, MnRI, University of Minnesota, Minneapolis, MN\quad
$^{2}$Michigan Tech\quad $^{5}$Hitech Kids}\\
{\small\tt \{mukhe100, hende471, tejpa005, mulla006, mcnei006, papan001, junaed\}@umn.edu,}\\
{\small\tt ajrhockey88@gmail.com, kris@hightechkids.org}
}
\date{}

\begin{document}

\maketitle

\blfootnote{\textcopyright~2026 IEEE. Personal use of this material is permitted. Permission from IEEE must be obtained for all other uses, in any current or future media, including reprinting/republishing this material for advertising or promotional purposes, creating new collective works, for resale or redistribution to servers or lists, or reuse of any copyrighted component of this work in other works.}
\blfootnote{The Institutional Review Board at the University of Minnesota reviewed this research under study number STUDY00024698 and determined it to be exempt from Human Subject Research. This research was supported by the National Science Foundation (NSF) \#IIS-2220956 and Minnesota Environment and Natural Resources Trust Fund (ENRTF) 2024-250, as recommended by the Legislative-Citizen Commission on Minnesota.}

\begin{abstract}

Engaging K-12 students in authentic scientific research remains a significant challenge, particularly at the intersection of environmental science and robotics. 
We introduce the Jar Jar ROV, a low-cost, open-source Remotely Operated Vehicle (ROV) platform designed for citizen science-based water quality monitoring by middle school students. 
This paper presents the design of the platform and the results of a large-scale deployment with over $100$ students across a US state who built, programmed, and deployed the ROVs in local lakes. 
The educational framework yielded high student engagement in hands-on activities, with ROV construction earning a perfect average score from mentors. 
From a scientific standpoint, the program successfully established a grassroots monitoring network, generating nearly eleven thousand validated measurements of temperature, pH, dissolved oxygen, and turbidity. 
However, our evaluation identified a critical ``engagement gap,'' with student interest declining sharply during more complex tasks such as electronics assembly and data uploading. 
This paper contributes both a validated, scalable model for integrating robotics into environmental education and a clear, data-driven roadmap for future improvements. 
These enhancements focus on lowering technical barriers and creating a more intuitive link between data collection and scientific discovery, addressing a key challenge in empowering the next generation of citizen scientists.


\end{abstract}
\section{Introduction}

Sustaining the health of our freshwater ecosystems requires widespread, consistent water quality monitoring, yet traditional methods are often too costly and resource-intensive for broad deployment~\cite{CHAPMAN2022132}. Many public access water bodies, such as rivers and lakes, contain pollutants and pathogens that can be harmful to humans, including bacteria like E. coli, chemical contaminants from agricultural runoff, and heavy metals from industrial sources \cite{MPCA2025_Ecoli} \cite{MPCA2025_HeavyMetals}. Beyond health concerns, other properties such as water clarity, temperature, dissolved oxygen levels, and pH are useful to know beforehand when planning underwater activities such as diving, swimming, or recreational boating. These parameters also serve as critical indicators of aquatic ecosystem health, affecting fish populations, plant growth, and overall biodiversity.
Concurrently, there are national imperatives to foster STEM engagement among K-12 students, particularly in the critical middle school years when interest in science and technology can either flourish or wane. 
While many educational initiatives exist, they often struggle to connect abstract concepts to tangible, real-world impact, leaving students disengaged. The challenge lies in simultaneously gathering vital environmental data and inspiring  the next generation of scientists and engineers.

\begin{figure}[htbp]
    \centering
    {
        \includegraphics[width=0.45\columnwidth, height=5cm]{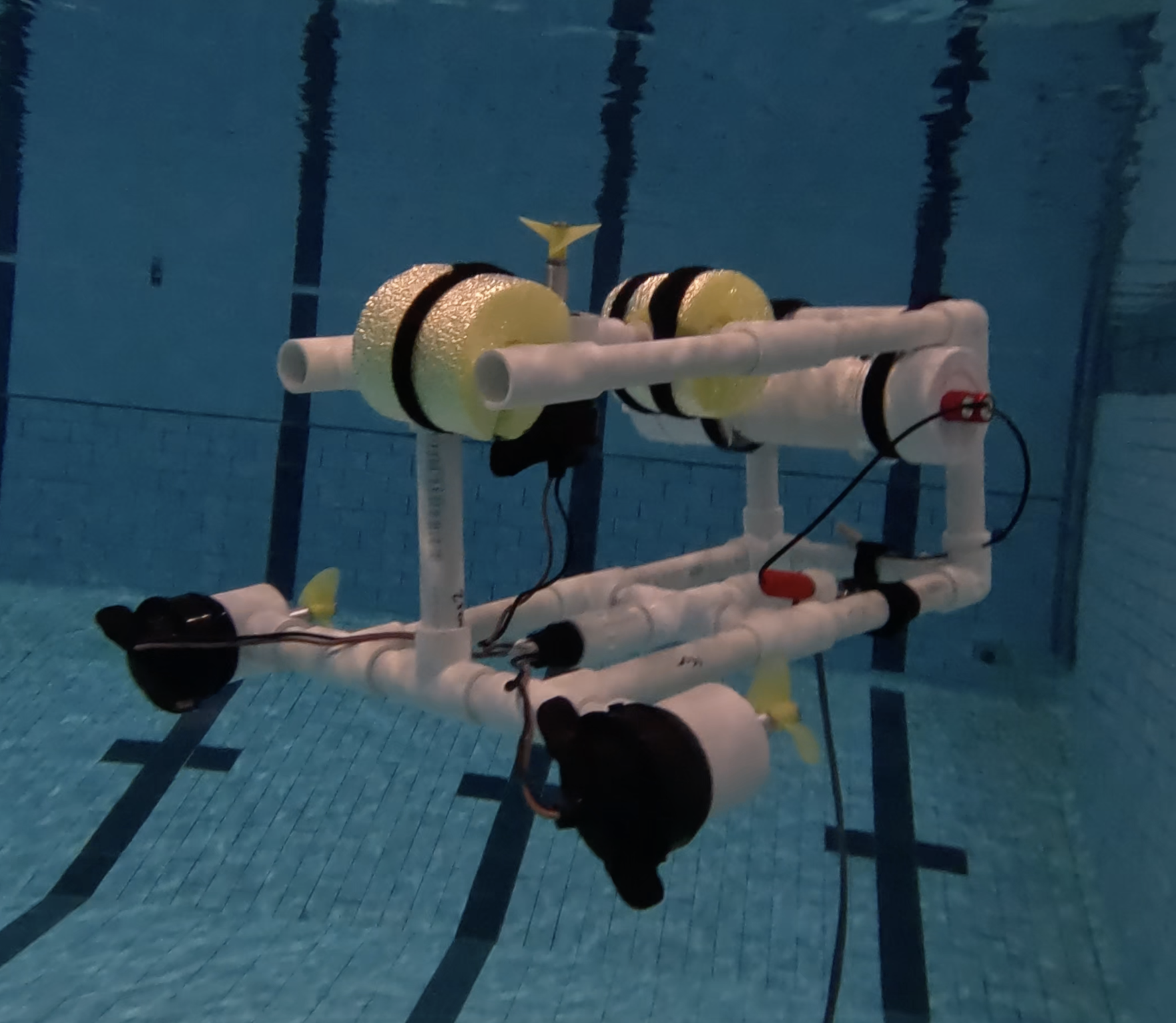}
        \label{fig:image111}
    }
    \hfill
    {
        \includegraphics[width=0.45\columnwidth, height=5cm]{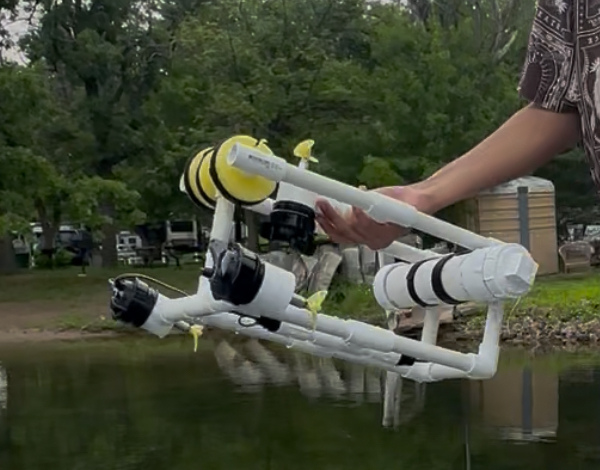}
        \label{fig:image222}
    }
    \caption{The Jar Jar ROV in a pool (left) and lake (right) with the attached water quality sensor pod. With the three-thruster configuration, the ROV can move in three axes, left-right (x/yaw), up-down (y/heave), and front-back (z/surge).}
    \label{fig:field-testing}
    \vspace{-1mm}
\end{figure}

In this paper, we argue that educational robotics offers a powerful and compelling solution to bridge this gap. 
A hands-on, purpose-driven robotics project can transform the abstract principles of engineering and environmental science into an exciting, goal-oriented mission. 
For a middle school student~\cite{Hussein}, the challenge is no longer just to learn about pH levels or sensor calibration in a classroom; it is to build and command a robot to explore the unseen world in their local lake. 
This shift in perspective is profound: the robot becomes the educational conduit -- a tangible vehicle that carries concepts of engineering, programming, and design directly into the service of scientific inquiry and environmental stewardship.

By empowering students to build and deploy their own Remotely Operated Vehicles (ROVs), we are not merely teaching them technical skills.
We are initiating them into the practice of citizen science ~\cite{6665185}, where their work contributes to a larger, authentic scientific endeavor. 
This approach addresses two fundamental challenges in STEM education: it provides a compelling ``why'' for learning difficult subjects and demonstrates that science is not just a body of knowledge, but an active, participatory process of discovery. 
The data they collect is no longer a set of numbers in a textbook; it is a direct measurement of the health of their own community's environment.

We thus present the Jar Jar ROV (Fig.~\ref{fig:field-testing}), a low-cost, open-source underwater robotics platform explicitly designed to serve this dual mission. 
This novel ROV is at a much lower price point than most commercially available platforms, less complex to operate, and more configurable than consumer-grade devices~(\eg \cite{chasing_dory} \cite{geneinno_titan_used_bhphotovideo}). 
Such properties significantly lower the barriers of entry to robotics and underwater robotics as a whole, for citizen scientists and K-12 robotics enthusiasts. 
We detail the development of the ROV and an accompanying curriculum, and report on a large-scale deployment with over $100$ students across a US state. 
Our findings demonstrate the immense potential of this model to generate both high levels of student engagement and a scientifically valuable dataset of freshwater quality. 
Furthermore, we analyze the key challenges encountered, offering a data-driven perspective on how to refine this approach to support K-12 citizen science at scale better.

\section{Related Work}
\label{sec:related}


\subsection{Educational Robotics} 
\label{sec:educational_robotics}

Educational Robotics has long been seen as an effective way to engage students in computational thinking, problem-solving, and STEM concepts \cite{eguchi2014}. 
Previous research indicates that robots can motivate young people to explore computer science and engineering, with case studies demonstrating their use in introductory programming courses \cite{9018567}.
Also, remotely accessible robotics platforms have been suggested to broaden research and learning opportunities at larger scales \cite{8793976}. 
Systematic reviews support these findings, showing that physical educational robots help develop skills like sequencing, debugging, and collaboration in young learners \cite{ching2023educational}. 
Building on this foundation, researchers have examined educational robots as autonomous systems that emphasize design, performance, and interaction. 
Comprehensive reviews have covered their effectiveness, motivation, and inclusion in robotics education \cite{anwar2019systematic}, while recent work illustrates how educational robots are evolving from classroom tools to platforms for innovation and citizen science \cite{lu2025decade}. 
Quantitative reviews demonstrate moderate positive impacts of educational robots on student performance, though results differ based on instructional methods \cite{ouyang2024effects}, emphasizing the crucial roles of pedagogy (the manner of instruction delivery) and robot embodiment in influencing educational outcomes \cite{huang2024rise}.
Large-scale programs such as FIRST\textregistered{} Robotics Competition (FRC) have shown significant positive impacts on students' STEM career choices and skills, with longitudinal studies demonstrating that participants are substantially more likely to major in engineering or computer science \cite{first_impact_2023} \cite{melchior_more_than_robots_2005}. 
However, the high cost, competitive focus, and steep learning curve create barriers to entry, highlighting the need for more accessible, curriculum-focused initiatives that prioritize scientific inquiry over competition.

\subsection{Citizen Science Initiatives}
\label{sec:citizen_science}

Citizen science offers possibilities to lower the barriers to entry for educational robotics, by allowing members with little expertise to participate in large-scale research through collaboration with high-expertise peers.
Technology-mediated citizen science projects have demonstrated success in strengthening global scientific infrastructure through distributed environmental monitoring, where volunteers use accessible technologies like Geo-Wiki and analyze data that would be infeasible for experts to gather alone \cite{6665185} \cite{FRITZ2012110}.
The integration of mechatronics into citizen science has enabled new forms of participatory sensing through low-cost mobile surface vehicles for environmental monitoring \cite{6665185} and community-based monitoring initiatives that engage local stakeholders in data collection \cite{danielsen_concept_2021} \cite{bonney_next_steps_2014}. 
Participatory science methods for water quality, combining volunteer-collected samples with remote sensing data, have proven effective in monitoring vulnerable near-shore habitats \cite{jordan_participatory_science_2023}.
However, these initiatives often rely on pre-built platforms or simple data collection tools, offering limited opportunities for participants to engage with the engineering and design process itself.

\subsection{Citizen Science \& Underwater Robotics}
\label{sec:usability_gaps}
Recent advancements in underwater robotics reveal a gap \donestarnote{JS: what kind of gap are we talking about?} in the low-cost, low barrier to entry ROV space. Commercial ROVs like the Chasing Dory (below \$1000) and Geneinno Titan (\$1000+) \cite{chasing_dory} \cite{geneinno_titan_used_bhphotovideo} exist, but they are not open-source or easily adaptable without domain-specific expertise. Open-source options such as the LoCO AUV and OpenROV require even more domain-specific expertise \cite{loco} \cite{open-rov} to build and operate, as well as support from maintainers in the case of issues. 
While some projects have developed low-cost underwater platforms for education, they often lack direct connection to citizen science applications \cite{open-rov}.
Existing water quality monitoring systems employ multiple methods;
\eg a community-based data collection approach, satellite-based data collection to measure water clarity \cite{umn_lakebrowser}, or manual sample collection~\cite{minneapolis_parks_water_monitoring}. 
These disparate approaches can result in sparse area coverage and many missing samples.
While some of these programs successfully engage volunteers in data collection, they face significant limitations in terms of the scope and precision of the measurements. 
They provide platforms for submitting data, but no clear guidelines on how to gather it. 
Additional inter-site challenges include a lack of cumulative temperature data, absent pH measurements, unreliable turbidity assessments (subjective Secchi disk readings rather than quantitative sensors), and unmonitored dissolved oxygen content as seen in \cite{minneapolis_parks_water_monitoring} \cite{metro_council_lake_monitoring} \donestarnote{JS: Do we have any citations to back this claim up?}.

The Jar Jar\donestarnote{JS: JAR JAR or Jar Jar or JarJar?} ROV project addresses these gaps by combining educational robotics with participatory citizen science through environmental monitoring, offering a solution that serves both as a cheap learning tool and a functional instrument for scientific contribution, filling the critical gap between high-cost/high-expertise systems and low-cost/low-functionality educational kits. 
This kit provides a two-fold benefit: (1) it allows volunteer citizen scientists to easily deploy their existing equipment with minor modifications, and (2) it leverages the scale of participatory science by operating as a yearly program where middle school youth gather data from local lakes across the state.

\section{Hardware \& Software Design}
\label{sec:ROV}

\subsection{ROV Design}
\label{sec:ROV_Design}

\begin{figure}[htbp]
    \centering
    \includegraphics[width=0.7\columnwidth]{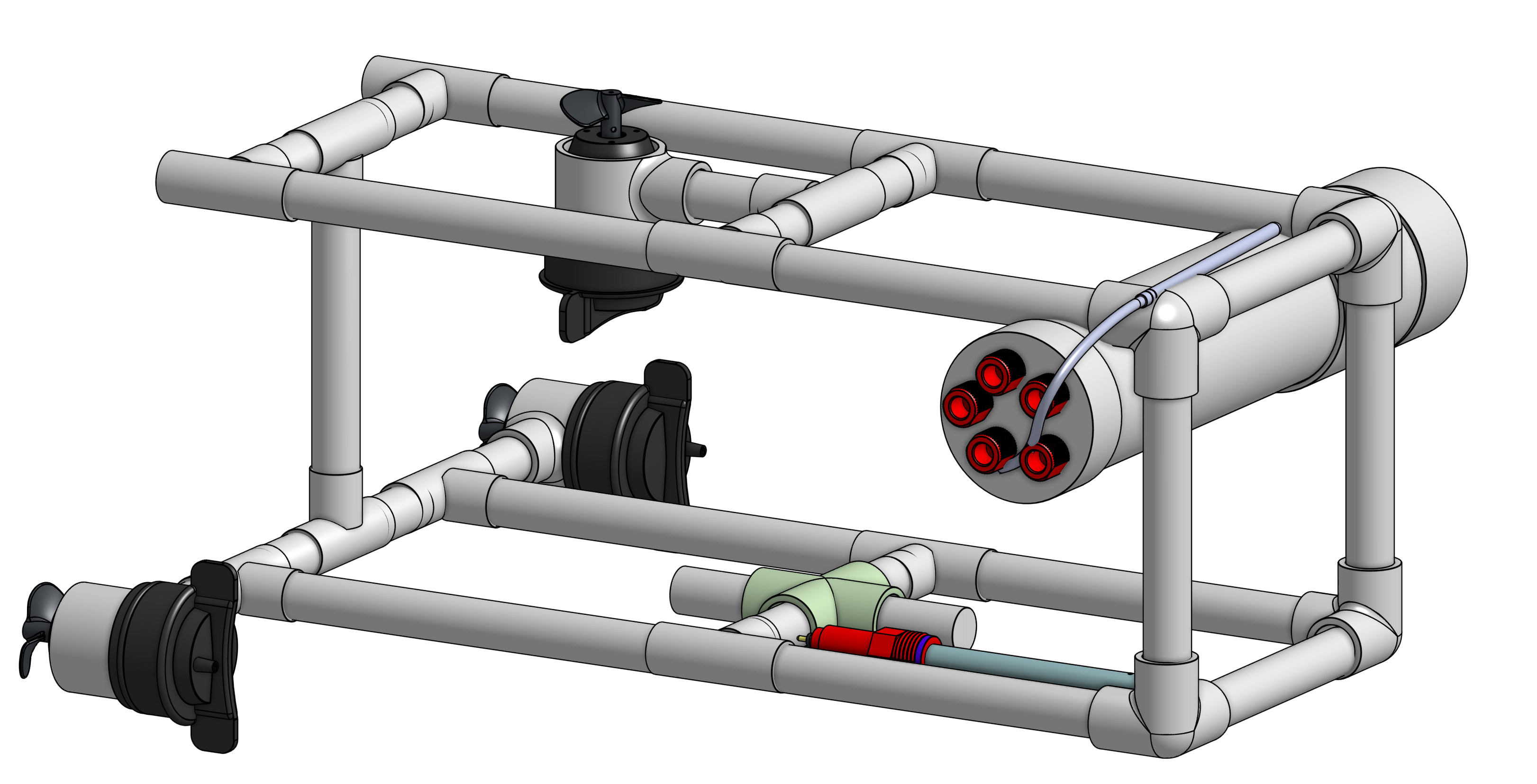}
    \caption{3D perspective of the ROV. The system features three bilge pump thrusters (black components), the cylindrical sensor pod (grayish white), and the tether connection point (green cross) for surface power delivery. The sensors are mounted on the frame.}
    \label{fig:frame}
    \vspace{-4mm}
\end{figure}

To build the ROV, we use an off-the-shelf propulsion and control system called the AngelFish ROV kit by SeaMATE \cite{seamate_angelfish2025}. 
This kit consists of a $12$V DC propulsion system ($3$ pump motors with propellers), a control system ($3$ switch control boxes for $3$-DOF locomotion), and complete wiring supplies, including a $10m$ \donestarnote{JS: No feet etc, meters in IEEE papers only} tether. 
To mount the propulsion system, we design a custom ROV frame made out of $15mm$ Schedule $40$ PVC pipes and $15mm$\donestarnote{JS:metric} PVC connectors. 
The frame design ensures adequate buoyancy control through the use of cylindrical foam attachments\donestarnote{JS: too informal: what are noodles made of? cylindrical foam? Can we just say that instead of ``pool noodles''?}, while maintaining stability for sensor deployment. 
With a material cost under $\$50$ and consisting entirely of PVC, the frame is extremely durable and easily repairable (Fig.~\ref{fig:frame}).

The frame geometry features a rectangular base structure measuring approximately $45cm\times20cm$ (length$\times$width) \donestarnote{JS: What does B stand for?} with vertical supports having a height of $18cm$ for thruster mounting and sensor pod attachment points. 
This configuration maintains the ROV's center of gravity below its center of buoyancy, ensuring stable operation even when carrying the sensor pod payload. 

\subsection{Sensor Pod}

\begin{figure}[htbp]
    \centering
    \includegraphics[width=0.8\columnwidth]{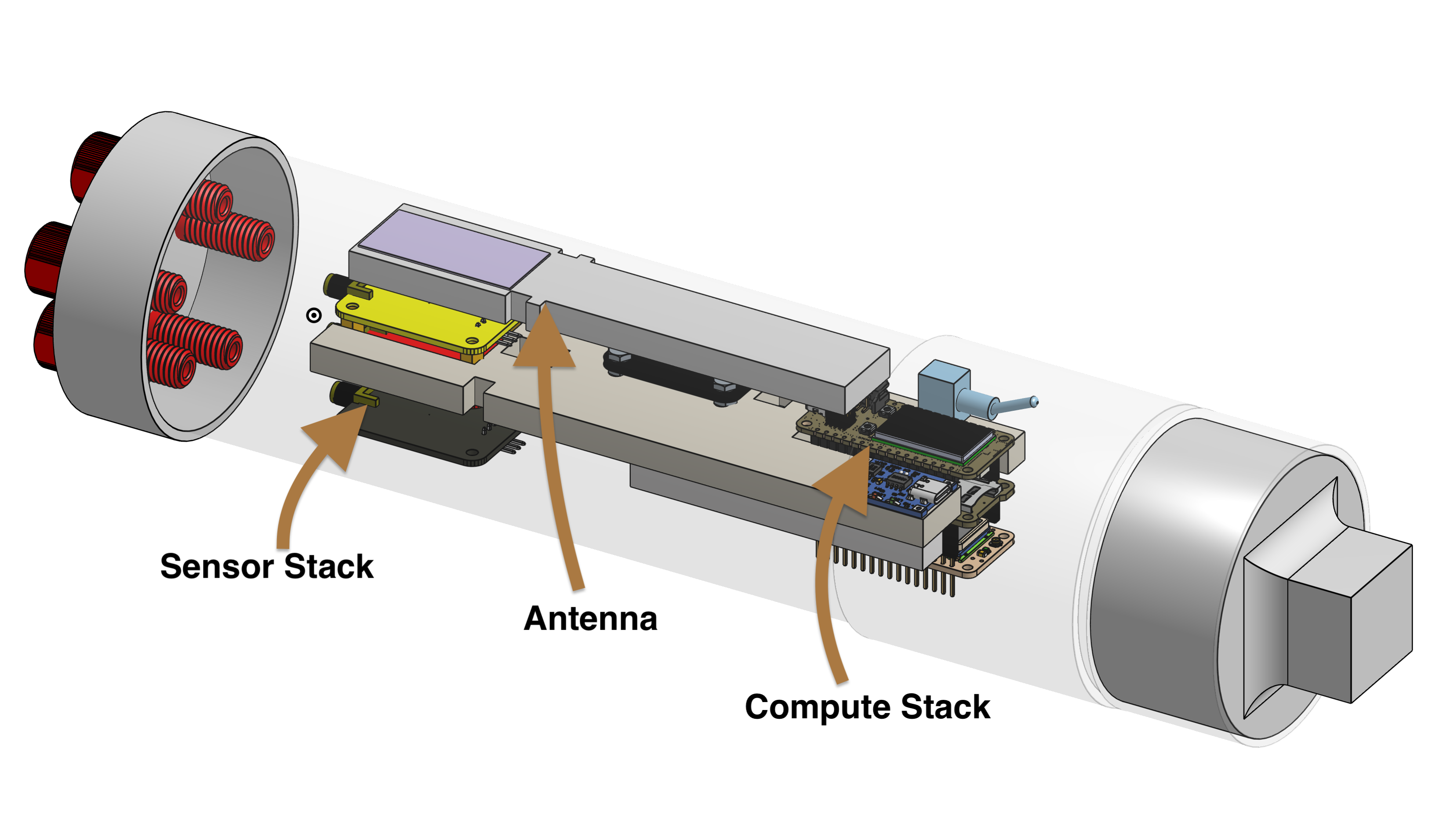}
    \caption{3D rendering of the sensor pod showing the placement of the internal electronics within the housing. Not pictures are the sensor probes that extend out of the red penetrators on the left side.}
    \label{fig:sensor-pod}
    \vspace{-4mm}
\end{figure}
The sensor pod, as shown in Fig.~\ref {fig:sensor-pod}, is entirely self-sufficient and does not require the ROV component to function. 
Its modular design allows it to be mounted on existing robotic platforms, surface vessels, or deployed as a standalone monitoring station. 
The primary goal of building the sensor pod is to create a versatile, cost-effective monitoring device that could easily be modified for different applications suitable for open-source citizen science initiatives, \donestarnote{JS: Figure 3 needs annotations: which is which? the reader has no way to know from this current figure.}

The mechanical design consists of a $230$ mm long, $60mm$\donestarnote{JS:metric} Schedule 40 PVC pipe chassis with $60mm$\donestarnote{JS:metric} PVC couplers and connectors. 
This diameter provides sufficient internal volume for electronics without causing the host ROV to be positively buoyant. 
The electronics are mounted on a custom MDF (Medium Density Fiberboard) board that slides into the tube, providing structural support and organized component layout. 
The apparatus achieves waterproof sealing with PVC cement on one end for permanent closure and a Teflon threaded PVC screw cap on the other end, maintaining a watertight seal tested at depths up to $10$ meters (when submerged for $1$ hour). 

\begin{figure}[htbp]
    \centering
    \includegraphics[width=1\columnwidth]{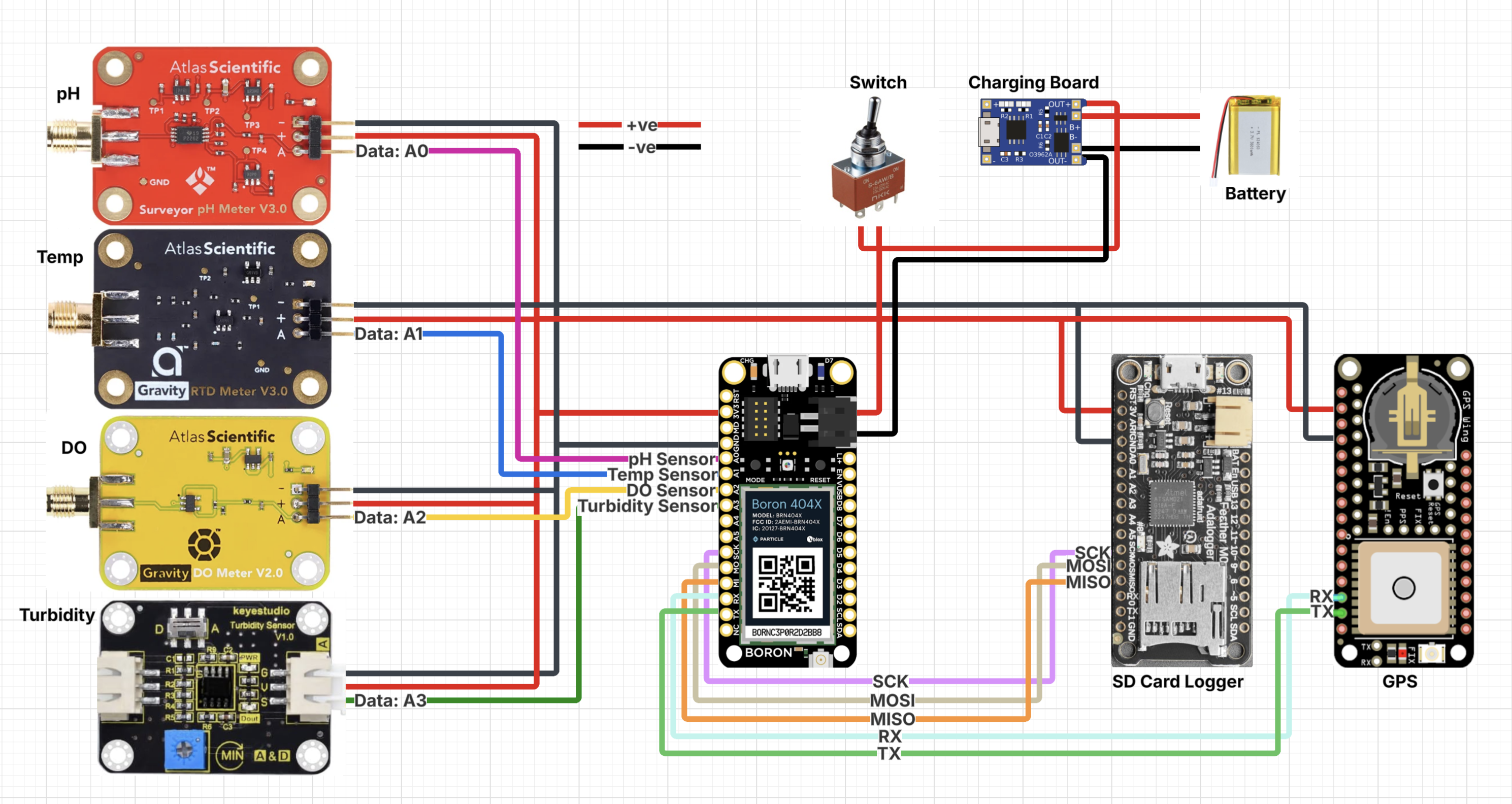}
    \caption{Water quality monitoring sensor pod circuit diagram showing Atlas Scientific pH, temperature, dissolved oxygen, and turbidity sensors connected to a Boron 404X cellular microcontroller with SD card logging, GPS positioning, and battery power management.}
    \label{fig:circuit-diagram}
    \vspace{-4mm}
\end{figure}

For the sensor electronics (Fig.~\ref{fig:circuit-diagram}), we use standard components off-the-shelf (COTS) that are readily available. 
The sensor pod microcontroller is a Particle Boron device that conforms to the `Featherwing' footprint \cite{adafruit_featherwing} \donestarnote{JS: what is featherwing? need citation.} $(2.29cm\times 5.08cm)$\donestarnote{JS:metric}, which is small enough to fit within the housing. 

\subsubsection{Computing Stack}
\label{sec:computing}
The computing stack consists of a Particle Boron microcontroller, an Adafruit Ultimate GPS-wing, and an Adafruit SD-logger wing. Component selection is driven by two\donestarnote{JS: I only see two items below?} major requirements: 

\begin{itemize}

    \item \textbf{Physical Requirements}: The main sensor pod housing is a $60mm$ diameter PVC tube, with limited space availability, eliminating platforms such as Arduino Unos, Raspberry Pis.

    \item \textbf{Complexity Management}: The Particle Boron streamlines setup with guided procedures via its console, reducing technical barriers for middle school students and educators. 
    Its built-in LTE enables direct cloud connectivity, unlike platforms such as the Raspberry Pi Zero or ESP32, which require additional cellular hardware and complex networking. 
    The Particle Cloud provides a web-based interface accessible on any device, allowing remote monitoring, firmware updates, and real-time data visualization without additional software. 
    It also supports device management and secure collaboration for educators.
    \donestarnote{JS: these two points seem very disjointed. What are you trying to say here? One is a constraint, but the other seems like a good thing, no? Aren't these better phrased as requirements than constraints?}
    
\end{itemize}

\subsubsection{Power System}
\label{sec:power_system}

The power system uses a $3600$ mAh battery capable of running the system for roughly $4$ hours (measured) of continuous operation. 
The system incorporates a module for safe battery charging, along with a power switch for user control of system activation. 
Power regulation circuitry within the Boron converts the battery voltage to the standardized $3.3$V operating voltage.
With the Boron thus serving as a central power distribution point, we eliminate the need for additional power conversion circuits and reduce overall system complexity.

\subsubsection{Sensor Suite}
\label{subsec:sen-suite}
The sensor suite comprises four integrated water quality sensors that measure the identified parameters: \textit{pH}, \textit{temperature}, \textit{turbidity}, and \textit{oxygen level}. 
The sensors interface with the Boron microcontroller through 12-bit analog-to-digital converters, providing sufficient resolution. 
pH calibration utilizes certified buffer solutions (pH $4.00$, $7.00$, $10.00$), and dissolved oxygen calibration involves air and control oxygen solutions. 
Temperature sensors undergo two-point calibration using an ice-water mixture ($0^o$C) and boiling water ($100^o$C at local atmospheric pressure). 
Lastly, turbidity calibration uses formazin standards prepared according to EPA Method 180.1 \cite{epa1993method180_1}. 
A modular architecture enables individual sensor replacement/calibration without any system-wide impacts.

\section{Experimental Design}
\label{sec:experimental_design}


\subsection{Overview}
\label{design_overview}
The experimental validation of our platform is conducted through a partnership with a local non-profit organization that coordinates educational activities for K-12 youth. 
The validation framework evaluates the educational effectiveness and technical performance of the system in a qualitative manner. 
The initial phase encompasses theoretical instruction where trained coaches deliver curriculum materials covering water quality fundamentals and ROV technology concepts. 
The subsequent deployment phase involves coordinated field testing at multiple lake sites across a US State, with teams systematically collecting water quality data using the constructed ROV systems. 
Lake selection prioritizes accessibility and educational value for this pilot implementation. 

\subsubsection{Parameter Selection}
\label{parameter_selection}
In this experiment, we select the water quality parameters as mentioned in~\ref{subsec:sen-suite} -- temperature, pH, oxygen content, and turbidity. 
These parameters are strategically selected for their conceptual accessibility to the target demographic while maintaining relevance to conservation efforts~\cite{waterSourcebook2007}. 
The \textit{technical goal} is for the teams to learn necessary skills to conduct a `lake trial' with the ROV kit.
The \textit{scientific goal} is for the teams to gather this data and compare their results with control values that were gathered earlier in the year by a team of experts.

\subsubsection{Educators \& Participants}
\label{educators_participants}
The water quality monitoring system targets middle school youth through a team-based approach. 
Each team comprises $4-7$ students supervised by two adult coaches, ensuring manageable group dynamics. 
The validation study encompasses $14$ out of $26$ registered teams involving over $100$ participants, being selected through an open application process administered jointly by our partners and parent institution. 
The selection methodology accepts all teams registering before the application deadline, maximizing program accessibility. However, due to scheduling conflicts, $12$ of the $26$ teams opted to complete the project in the following summer. 

\subsubsection{Educational Materials}
\label{educational_materials}
The instructional content is organized into two complementary modules addressing distinct learning objectives:
(1) The \textit{Water and Lake module} provides foundational knowledge of the $4$ key water quality parameters; 
(2) The \textit{Robotics module} delivers systematic technical instruction covering the complete construction workflow through detailed step-by-step procedures. 
The instructional sequence progresses logically from frame fabrication and assembly through sensor pod construction, electronics integration, and software configuration. 
Supplementary materials include calibration protocols and field deployment guidelines, ensuring consistent data collection methodology across all teams. 
Educational resources are maintained with open-source access through the project website. 
We also provide a YouTube\texttrademark{} channel that contains videos to complement the written materials.

The curriculum delivery or `lesson plans' spans over a structured three-month timeline. 
The first month focuses on water quality foundations. 
Subsequently, robotics building activities provide hands-on application of engineering principles. 
The final phase integrates theoretical knowledge with practical field experience through guided data collection activities. 
Upon successful field deployment, teams upload their measurements to a centralized database on the project website, constituting the citizen science component of the study. 
Within this experimental framework, distributed teams across the target US State are generating spatially comprehensive water quality datasets tracking lakes throughout the state. 
The overarching goal is to evaluate the feasibility of establishing grassroots infrastructure for systematic water quality gathering, specifically assessing data quality and educational accessibility.

\section{Training \& Field Testing}
\label{sec:training_field_testing}

\begin{figure}[htbp]
    \centering
    {
        \includegraphics[width=0.45\columnwidth, height=5cm]{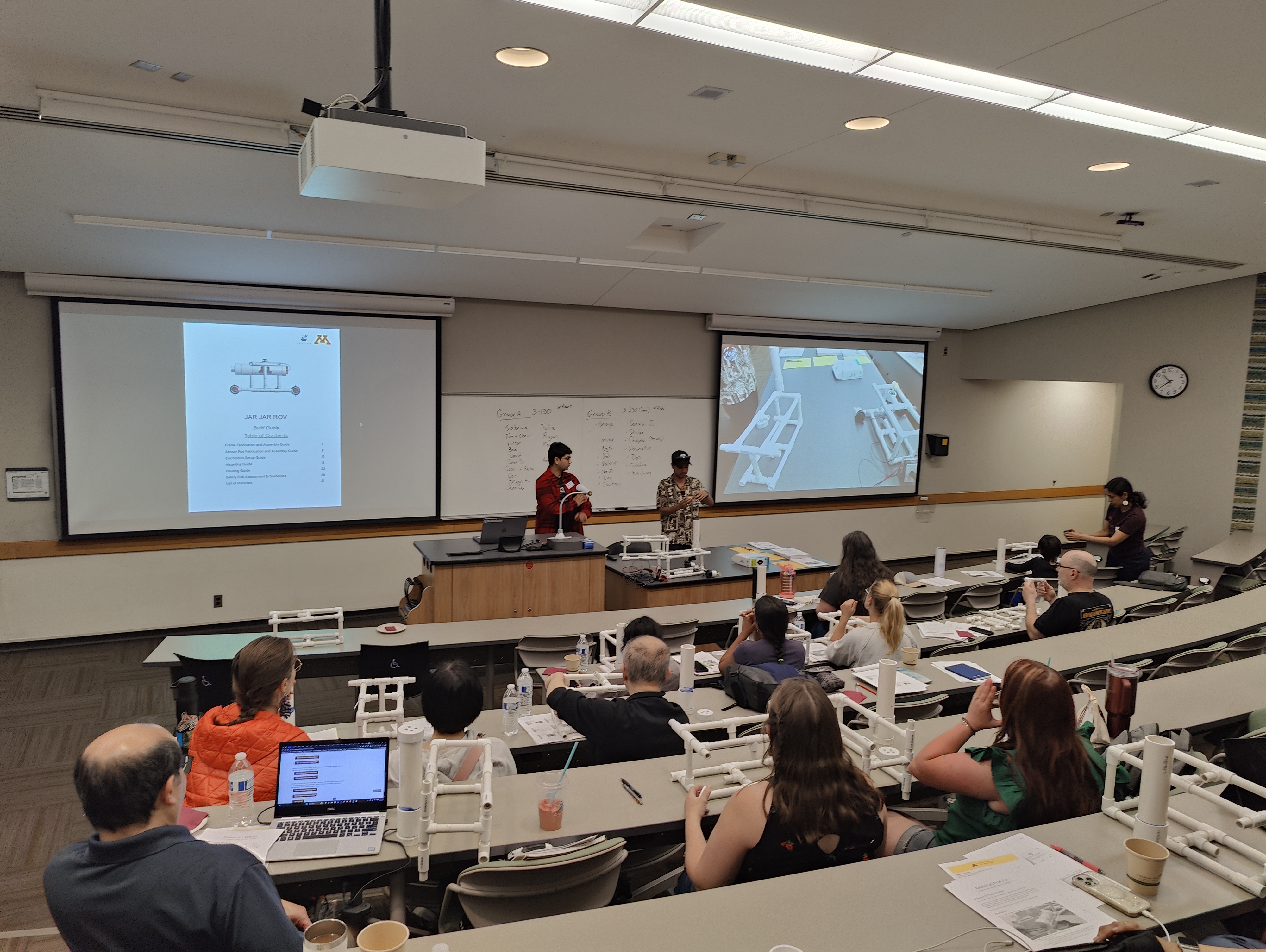}
        \label{fig:image111}
    }
    \hfill
    {
        \includegraphics[width=0.45\columnwidth, height=5cm]{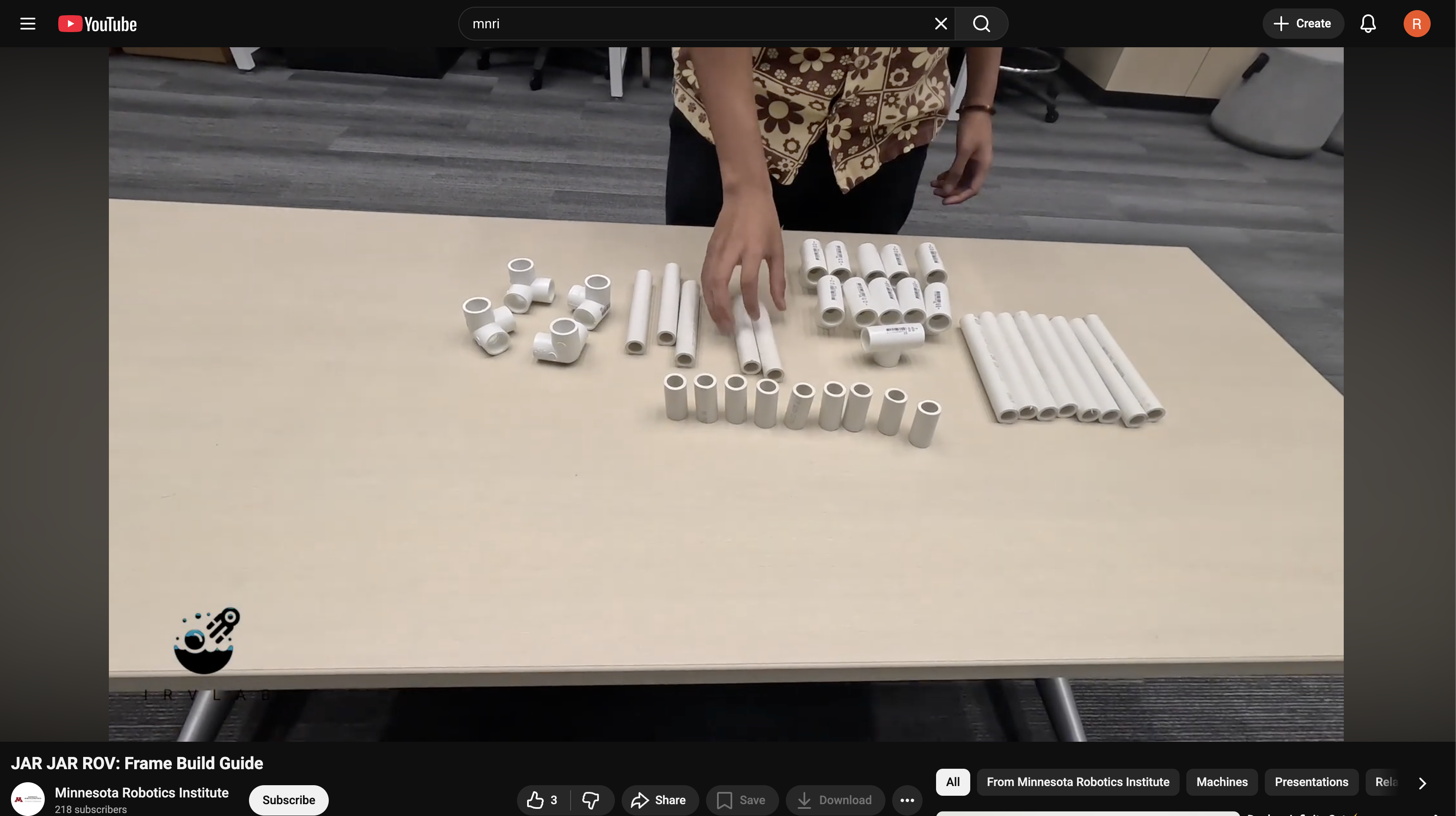}
        \label{fig:image222}
    }
    \caption{On the left is an image of the in-person training where teams are taught skills required to build, operate, and maintain the kit. On the right is an image of the recorded online training where teams are taught other relevant skills. }
    \label{fig:field-training}
\end{figure}

\donestarnote{Figure 5 (right) remove faces and logos!}
The field testing program employs a two-part training approach (as shown in Fig.~\ref{fig:field-training}) designed to ensure technical competency while maintaining accessibility for the target demographic. 
The first part consists of an in-person training workshop for educators before the program begins. 
Educators construct the ROV kit and undergo training for assembling the electrical components. 
The session includes comprehensive software training, covering the Particle Boron microcontroller architecture, Atlas Scientific sensor suite operation, and critical deployment protocols, including ballasting techniques for achieving neutral buoyancy and tether management procedures. 
The second part is an online programming workshop to provide further technical instruction in microcontroller programming, including Particle device configuration, cloud connectivity establishment, and sensor calibration procedures.

\begin{figure}[htbp]
    \centering
    \includegraphics[width=1.0\columnwidth]{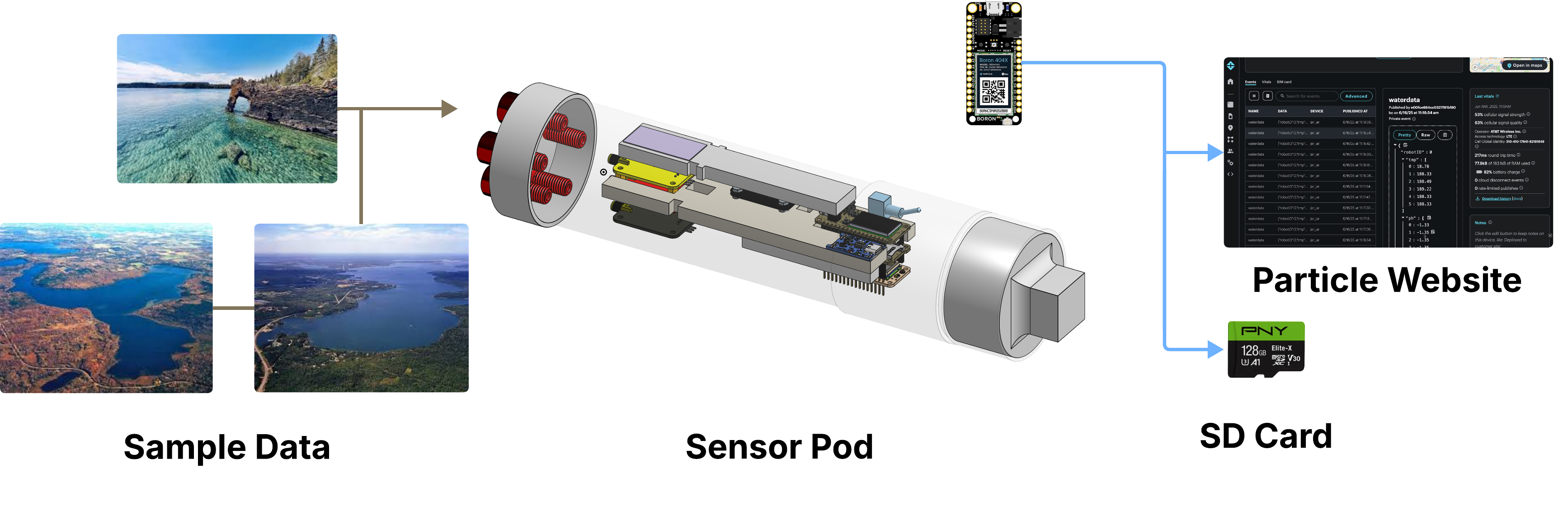}
    \caption{A visual representation of the data collection pipeline in the sensor pod. The raw data is collected in two second intervals on the SD card. The website receives the cumulative set of values every ten seconds when able. }
    \label{fig:flow-chart}
    \vspace{-4mm}
\end{figure}

Teams are assigned to specific lake sites and follow a standardized deployment protocol beginning with systematic pre-deployment equipment testing. 
Field deployment operations require teams to conduct $30$-minute data collection sessions, including site safety assessment, system initialization with GPS positioning confirmation, and systematic underwater surveys covering multiple depth levels. 
Teams collect continuous measurements of pH, temperature, dissolved oxygen and turbidity throughout $30$ minutes, following the pipeline of Fig.~\ref{fig:flow-chart}. 
Following completion, teams upload collected datasets to a centralized repository, which undergoes automated and manual cleaning and is incorporated into the project website weekly. 
The teams present their findings at the local state fair where they are judged on their work, concluding the experiment.

Post-assessment surveys indicated substantial knowledge gains in both environmental science and robotics concepts, with participants showing an increased interest in STEM. 
The structured program successfully validated both the technical feasibility of student-operated environmental monitoring systems and the educational effectiveness of authentic scientific participation through citizen science methodologies.
\section{Results}
\label{sec:results}

The results and observations from the pilot program's deployment of the Jar Jar ROV are described below, along with a data-driven analysis of the student-collected measurements and qualitative findings from mentor feedback. 
We identify key areas of improvement, such as the low data retention rate and the ``engagement gap'' in technical tasks.

\subsection{Educational Outcomes}
The project utilizes a post-program survey, shown below in its entirety, sent to team mentors to assess the program's viability qualitatively. 
The survey also contains technical questions aimed at the logistics of the program to help adapt it for future iterations. 
The responses to these questions do not directly provide insight into the results of the program; instead, they are addressed in Sec.~\ref{sec:Improvements}.

\begin{figure}[t!]
    \centering
    
    \begin{subfigure}{0.95\columnwidth}
        \centering
        \includegraphics[width=\textwidth]{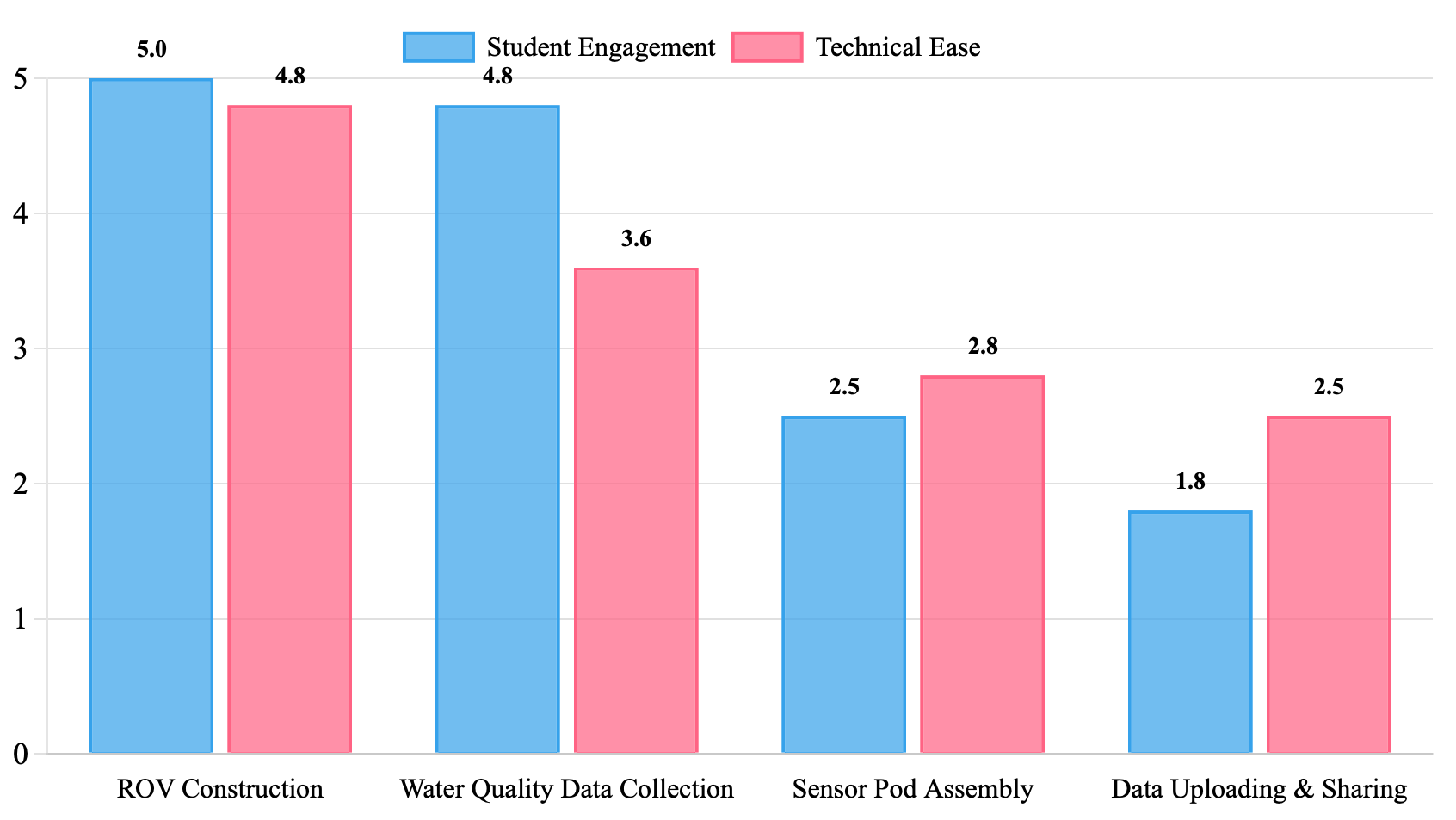}
    \end{subfigure}
    
    \caption{Survey responses indicate most interest and accessibility in the ROV construction phase, with the field experiment section involving water quality data collection at a close second. The sensor pod assembly and data upload phases were less engaging. Technical ease seemed to follow the student engagement. Higher values indicate favorability.}
    \label{fig:rov_program_results}
    \vspace{-0.5cm}
\end{figure}
\begin{center}
    {\textbf{Summary of Survey Questions}}
    \vspace{0.5em}
    \hrule
\end{center}

\subsection*{Mentor \& Team Background}
\begin{itemize}
    \item What is your role (e.g., coach, teacher)?
    \item Did you have prior experience with robotics or environmental science?
    \item How many students were on your team?
\end{itemize}

\subsection*{Training \& Preparation}
\begin{itemize}
    \item How clear and helpful were the training materials provided (manuals, guides, online resources)?
    \item Did you feel adequately prepared to mentor the project after the training?
    \item What suggestions do you have for improving the training materials?
\end{itemize}

\subsection*{ROV Kit \& Building Process}
\begin{itemize}
    \item How easy was it for your team to build the ROV frame (PVC structure)?
    \item How easy was it to assemble and calibrate the sensor pod?
    \item How easy was it to integrate the electronics and wiring?
    \item How easy was the field deployment?
    \item What challenges did you encounter with the kit?
\end{itemize}

\subsection*{Curriculum \& Water Quality Booklet}
\begin{itemize}
    \item Did the curriculum modules align well with student comprehension levels?
    \item How often did your team use the Water Quality Activity Booklet?
    \item Which booklet activities were most engaging for students?
    \item How did the booklet's engagement level compare to the robotics build?
    \item Did the booklet help you explain water quality concepts (e.g., pH, turbidity)?
    \item Did the booklet increase students' understanding of local environmental issues?
\end{itemize}

\subsection*{Student Outcomes}
\begin{itemize}
    \item How engaged were students during different phases (ROV construction, sensor pod assembly, data collection, data uploading)?
    \item In your opinion, how much did students learn about water quality science, robotics, teamwork, and citizen science?
    \item Did you notice any change in your students' interest in STEM or environmental science?
\end{itemize}

\subsection*{Overall Feedback \& Future Participation}
\begin{itemize}
    \item What were the most rewarding and most challenging aspects of mentoring?
    \item How satisfied were you with your overall experience?
    \item Would you be interested in mentoring a similar project in the future?
\end{itemize}

\vspace{0.5em}
\hrule
\vspace{0.5em}

Eight of fourteen educator teams responded to the survey (before the deadline), with six teams indicating definitive interest in future participation and two teams expressing conditional interest. 
Students showed varying engagement across the program components (Fig.~\ref{fig:rov_program_results}): they achieved perfect engagement during ROV construction (mean=$5.0$) and high engagement during water quality data collection (mean=$4.8$), but showed markedly lower engagement with sensor pod assembly (mean=$2.5$) and data upload (mean=$1.8$). 
The small sample size ($n=8$) prevents the chi-squared test from providing any meaningful insights. 
The Pearson correlation test \cite{schober_correlation_2018}, on the other hand, revealed a highly significant positive relationship between technical ease and student engagement ($r=0.90$, $t=5.09$, $p<0.01$, $n=8$, where $r$ is the Pearson coefficient, $t$ represents the value of the T-test \cite{wadhwa_t_2023}, $p$ is the ``p value'' \cite{dahiru_p-value_2008}, and $n$ is the sample size), with $81\%$ variance in engagement explained by technical difficulty. 
This strong correlation indicates that for every one-point improvement in technical ease, student engagement increases by approximately $1.4$ points on the five-point scale.
Educators also reported students achieved the strongest knowledge gains in water quality science (mean=$4.4$), followed by citizen science applications (mean=$4.1$), teamwork (mean=$4.0$), and robotics and engineering (mean=$3.8$). 
All responding teams encountered implementation challenges, with seven teams reporting electronics-related difficulties, including sensor calibration issues, fragile wiring connections, and complex assembly procedures that exceeded student capabilities. 
These technical barriers constitute a significant pedagogical constraint that may impede program scalability. Despite implementation challenges, educators maintained moderate to high satisfaction levels, with $50\%$ reporting they felt ``very satisfied'' and $50\%$ ``somewhat satisfied''. While the program demonstrates educational promise, further refinement of the sensor pod design is needed to make it more accessible and reliable before being released.
\donestarnote{JS: need to report some statistical significance of the test results on the data -- \eg T-Test/ANOVA/Chi-Squared!}

\subsection{Citizen Science Impact}
\label{sec:citizen_science_impact}

\begin{figure}[thb]
    \centering
    \vspace{-1mm}
    \includegraphics[width=1.0\linewidth]{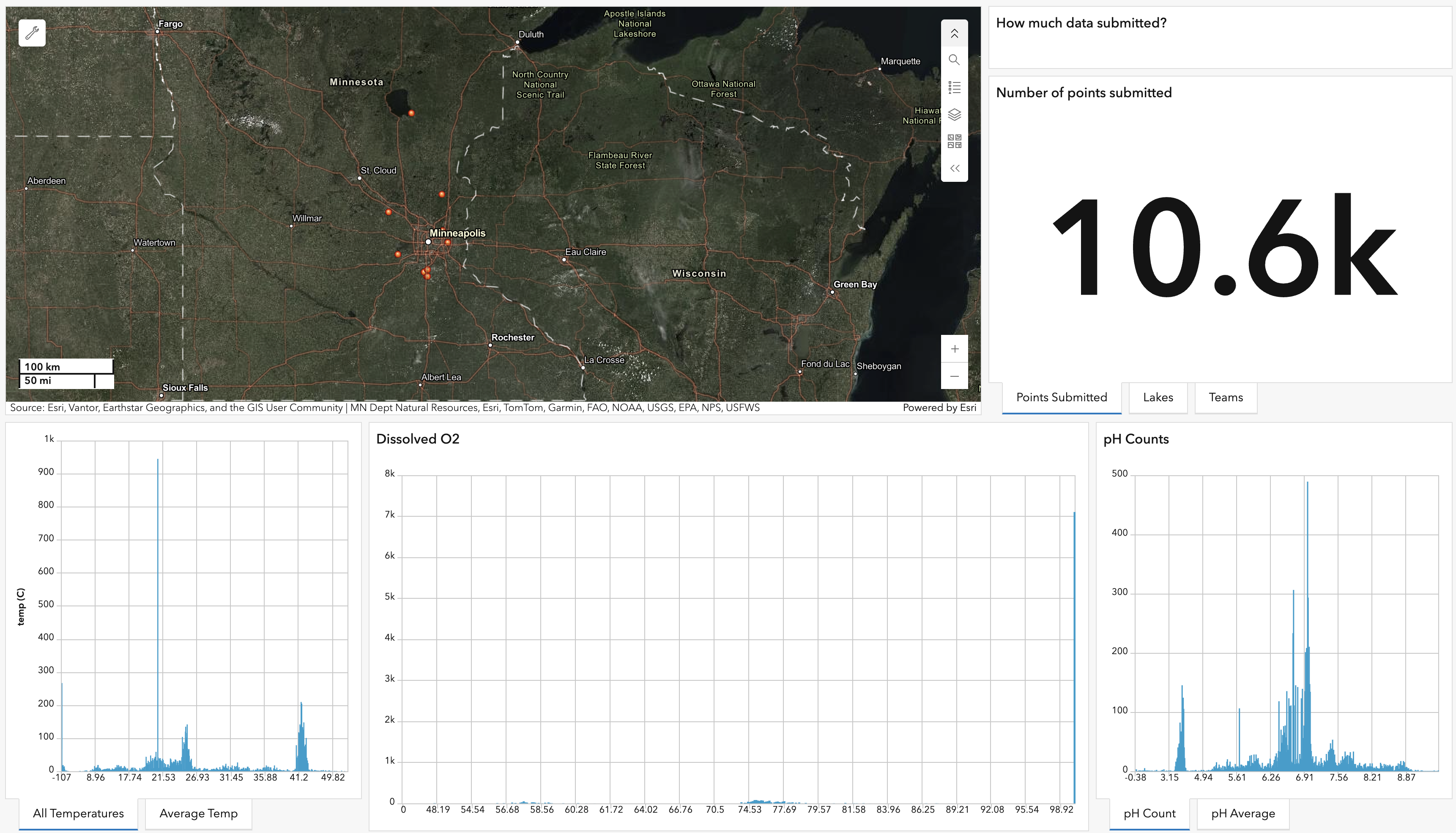}
    \caption{A screenshot of the interactive map website where the public can access collected water quality data. The red dots are the sample points taken at different lake sites.}
    \label{fig:map}
    \vspace{-0.5cm}
\end{figure}

From over $30,000$ raw sensor measurements, $10,600$ usable samples ($32.8\%$) are retained after quality control, as seen in Fig.~\ref{fig:map}\donestarnote{JS: map shows Minnesota, another violation of double-blind rules!}; the retention rate jumps to $100\%$ when samples are averaged into a singular daily value, \eg all temperature values are averaged to a single reading for the day.\donestarnote{JS:What does singular daily value mean?} 
Water quality data revealed characteristic summer conditions: $22^o$C average temperature, $48\%-99\%$ dissolved oxygen saturation, and bimodal pH distribution (with peaks at $6.9$ and $7.6$) consistent with regional lake data.
While satellite remote sensing systems from the parent institution provide valuable broad-scale monitoring through daily and monthly averaged measurements across entire lake surfaces, the Jar Jar ROV system offers complementary high-resolution, ground-truth measurements that capture more granular temporal and spatial variability missed by averaged satellite data. 

\begin{table}[htbp]
    \centering
    \small  
    \begin{tabular}{@{}l c c@{}}
    \toprule
    \textbf{ROV Platform} & \textbf{Cost} & \textbf{Academic Level} \\
    \midrule
    Chasing Dory & \$600 & Consumer \\
    Geneinno Titan & \$2k+ & Consumer \\
    OpenROV & \$899-1.5k & Undergrad+ \\
    LoCO AUV & \$2.5k+ & Undergrad+\\
    \textbf{Jar Jar ROV} & \textbf{\$300} & \textbf{Middle School} \\
    \bottomrule
    \end{tabular}
    \caption{Cost comparison of ROV some example open-source and commercial platforms.}
    \label{tab:rov-comparison}
    \vspace{-2mm}
\end{table}


\subsection{Cost-Effectiveness and Scalability}
\label{sec:cost_scalability}

Table~\ref{tab:rov-comparison} provides prices of a sample of commercially-available and open-source systems in comparison with that of the Jar Jar ROV, highlighting its significantly-cheaper cost.
The sensor kit, built using the most-affordable off-the-shelf components from reputable manufacturers, costs $\$290$. 
The total cost of the program, including costs for tools, falls at roughly $\$1000$ per team. 
Popular nation-wide existing programs in the US, such as~\cite{first_impact_2023}, cost over $\$6000$ per team, implying the cost of our program is one-sixth of the cost of existing platforms, making it more accessible and scalable. 
High mentor retention and standardized curriculum materials demonstrate significant scalability potential across K-12 educational environments. 
Additionally, the ROV kit is reusable, so teams can repeat the program on a yearly basis. 
Finally, the modular and affordable nature of the kit makes it an ideal tool for existing citizen scientists and volunteers to obtain scientific data. 
\section{Pilot Program Insights}
\label{sec:Improvements}

The initial year of the project serves as a pilot run, uncovering potential issues. 
The feedback gathered is therefore a critical component of our iterative design process. 
The following months, leading up to the program's repetition in May 2026, will be dedicated to a comprehensive enhancement phase. 
This period will be used to systematically address the identified issues to prepare the program for a permanent and scalable release. 
We use the survey and logs from the program to inform the improvements that need to be made.

\subsection{Hardware Reliability Analysis}
\label{sec:reliability_analysis}

Hardware reliability testing across $14$ deployed units revealed specific component vulnerabilities that require targeted improvements. 
Table~\ref{tab:hardware_reliability} summarizes the failure rates observed during the pilot deployment.

\begin{table}[htbp]
    \centering
    \begin{tabular}{@{}l c c c@{}}
    \toprule
    \textbf{Component} & \textbf{Failures} & \textbf{Total} & \textbf{Rate} \\
    \midrule
    Boron Microcontroller & 0 & 14 & 0\% \\
    Temperature Sensor & 0 & 14 & 0\% \\
    Dissolved Oxygen Sensor & 0 & 14 & 0\% \\
    Charging Board & 0 & 14 & 0\% \\
    pH Sensor & 1 & 14 & 7.1\% \\
    Turbidity Sensor (Original) & 8 & 14 & 57.1\% \\
    \textbf{Turbidity Sensor (Refined)} & \textbf{0} & \textbf{8} & \textbf{0\%} \\
    \bottomrule
    \end{tabular}
    \caption{Hardware Component Reliability Analysis across 14 deployed units during pilot testing. The turbidity sensor redesign with a refined seal eliminated all failures.}
    \label{tab:hardware_reliability}
    \vspace{-0.5cm}
\end{table}

There were two instances of `catastrophic failures' where the sensor pod itself had a leak, resulting in total loss of electronics. 
The cause of this was attributed to user error, as both teams missed the crucial step of sealing the Pod before submersion. 
The two teams were provided replacements, and with the correct seal, they did not suffer any further issues. 
The two catastrophic failures have thus been omitted from Table~\ref{tab:hardware_reliability}. 
The turbidity sensor emerged as the primary reliability concern, with a $57.1\%$ failure rate in the original design. 
However, a mid-deployment redesign incorporating an improved seal completely eliminated failures in subsequent units. 
Future hardware iterations will also implement simplified electrical connections that minimize or eliminate soldering requirements, making the project more accessible to educators with varying technical backgrounds, as informed by the survey. 
The sensor pod will be redesigned to address the ``very tight fit'' of the breadboard, and enhanced calibration guides will include comprehensive troubleshooting steps covering multiple common failure scenarios.

\subsection{Data Collection Performance}
The pilot program demonstrated exceptional data collection capabilities, gathering over $30,000$ samples across all teams. 
Of these, $10,600$ samples were retained for analysis after manual data cleaning processes. 
Each of the $14$ teams collected approximately $2400-3000$ samples per day, compared to baseline community monitoring methods that typically collect one sample per day per site (\eg\cite{umn_lakebrowser,metro_council_lake_monitoring})\donestarnote{JS: ok to leave these references, as you do not refer to them as ``our baseline'' or something similar}. 
The low $32.8\%$ sample retention rate is a result of teams not cleaning up their raw data files; when the sensor pod is powered, the device begins logging data until it is powered down. 
This results in intervals where data is being collected outside of the target environment, which need to be removed for data sanity. 
Additionally, other outliers that can be trivially spotted (\eg, unreasonably low or high water temperatures) need to be erased.
This manual cleanup procedure is basic enough for both educators and students to follow, resulting in higher sample retention and eliminating the need for cleanup on the administrative side. 
These instructions will be added to the educational materials, along with improved calibration and data validation guides to ensure data quality and consistency.

\subsection{Curriculum and Training Improvements}
Feedback indicates that training materials are at times disconnected and overwhelming, while some sections of the lake curriculum feel too academic for a summer program. 
To address these concerns, we will unify training materials with difficulty ratings and time estimates for each task, allowing our partners to structure the lesson plans better. 
The curriculum will be further simplified to be more engaging for middle school audiences, incorporating hands-on experiments and bite-sized instructional videos while creating clearer connections between ROV data collection and real-world scientific impact. 
These improvements are designed to improve engagement, reduce the barrier to entry, and improve accessibility for those outside the program as well.
\section{Conclusion}

This paper presents the design, development, and validation of the Jar Jar Remotely Operated Vehicle (ROV), a cost-effective platform designed to bridge the gap between educational robotics and authentic citizen science for water quality monitoring. 
Through a pilot deployment with middle school students, we have demonstrated the system's technical feasibility and educational effectiveness. 
It is an effective educational tool that yields substantial knowledge gains and high student engagement, particularly in hands-on activities like ROV construction and data collection. 
Furthermore, the project successfully managed a grassroots environmental monitoring network, with student teams collecting $10,600$ validated water quality measurements. 
The success of this pilot, evidenced by its low cost of $\$300$ per unit and positive feedback on future participation, validates the potential for establishing scalable, grassroots infrastructure for systematic environmental monitoring. 
The feedback gathered has provided a clear roadmap for future work, which will focus on creating a unified master guide, redesigning the hardware for greater robustness, and further refining the curriculum to enhance the user experience. 
Ultimately, the Jar Jar ROV project serves as a model for leveraging accessible technology to empower young minds as active contributors to scientific research, fostering the next generation of environmentally conscious citizens.

\section{Future Work \& Modifications}
\label{sec:future-work}


\begin{figure*}[!t]
    \centering
    \begin{subfigure}[b]{0.62\linewidth}
        \centering
        \includegraphics[width=\linewidth]{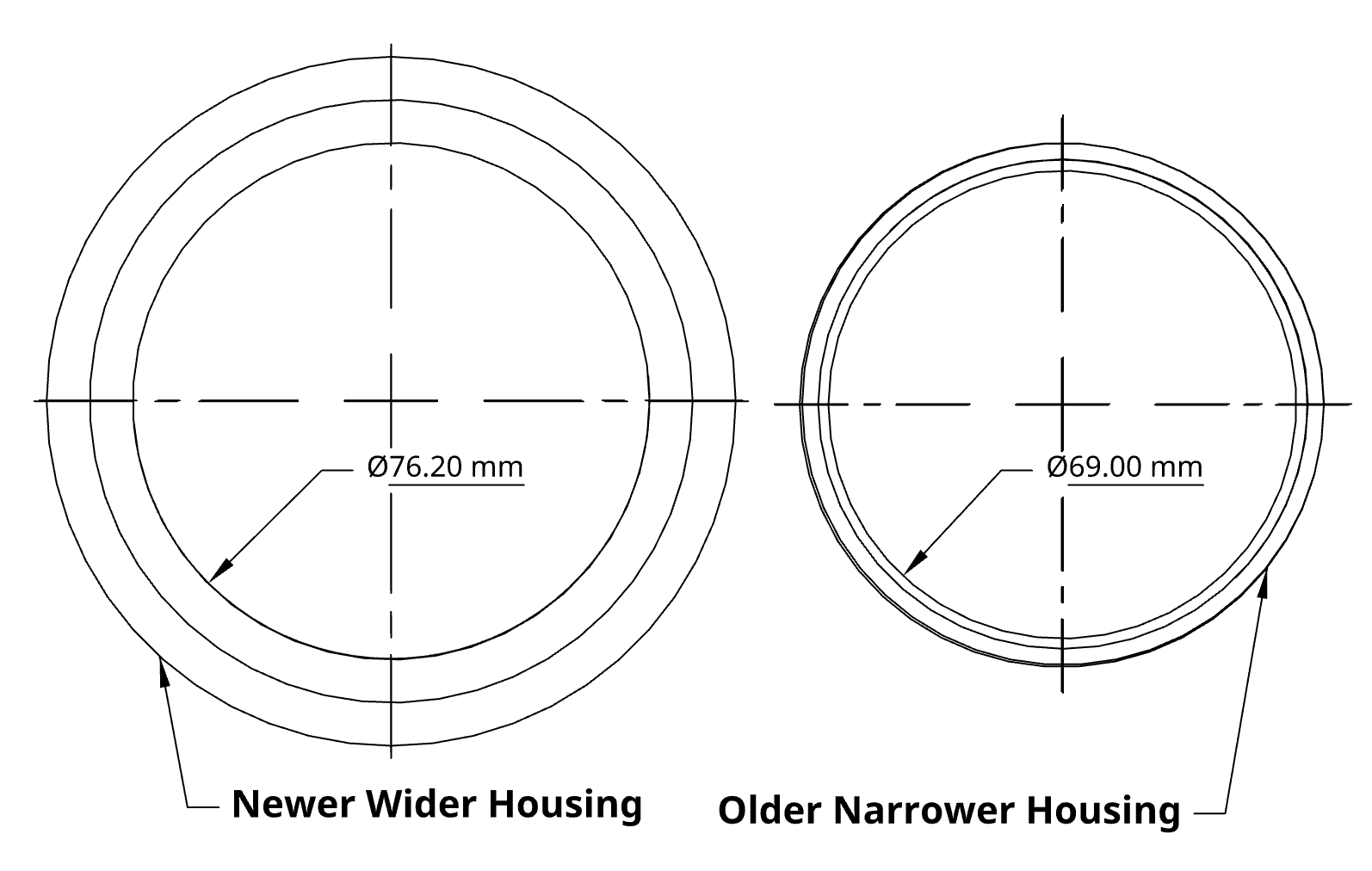}
        \caption{}
        \label{fig:revised-assembly-drawing}
    \end{subfigure}
    \hfill
    \begin{subfigure}[b]{0.36\linewidth}
        \centering
        \includegraphics[width=\linewidth]{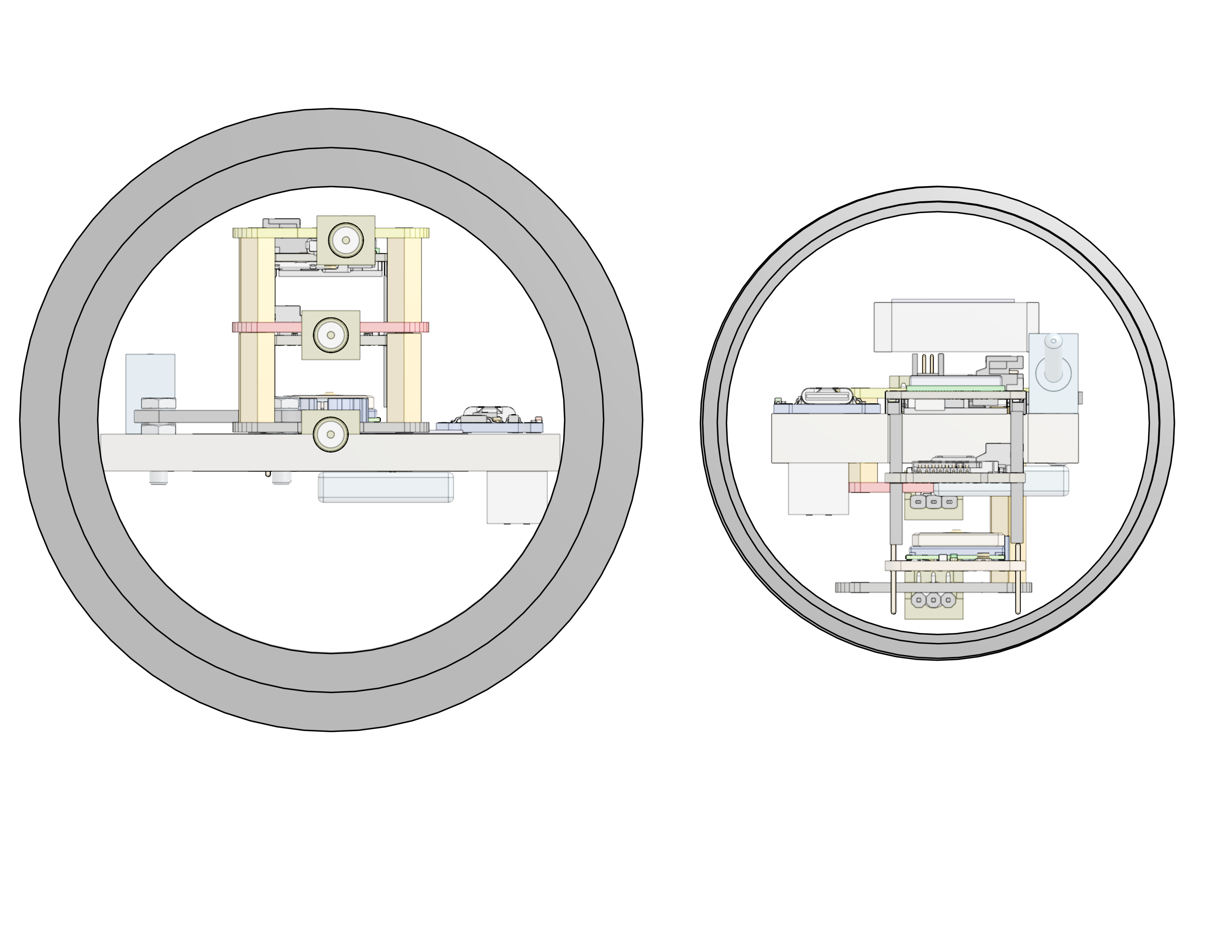}
        \caption{}
        \label{fig:revised-assembly}
    \end{subfigure}
    \caption{(a) shows the newer housing compared to the older housing with regard to internal volume. (b) shows how the internal electronics are housed in the modified design. The newer design provides more room to the wires and is held in place by the board sitting on the PVC wall.}
    \label{fig:row1}
\end{figure*}

\begin{figure*}[t]
    \centering
    \setcounter{subfigure}{2}
    \begin{subfigure}[b]{0.62\linewidth}
        \centering
        \includegraphics[width=\linewidth]{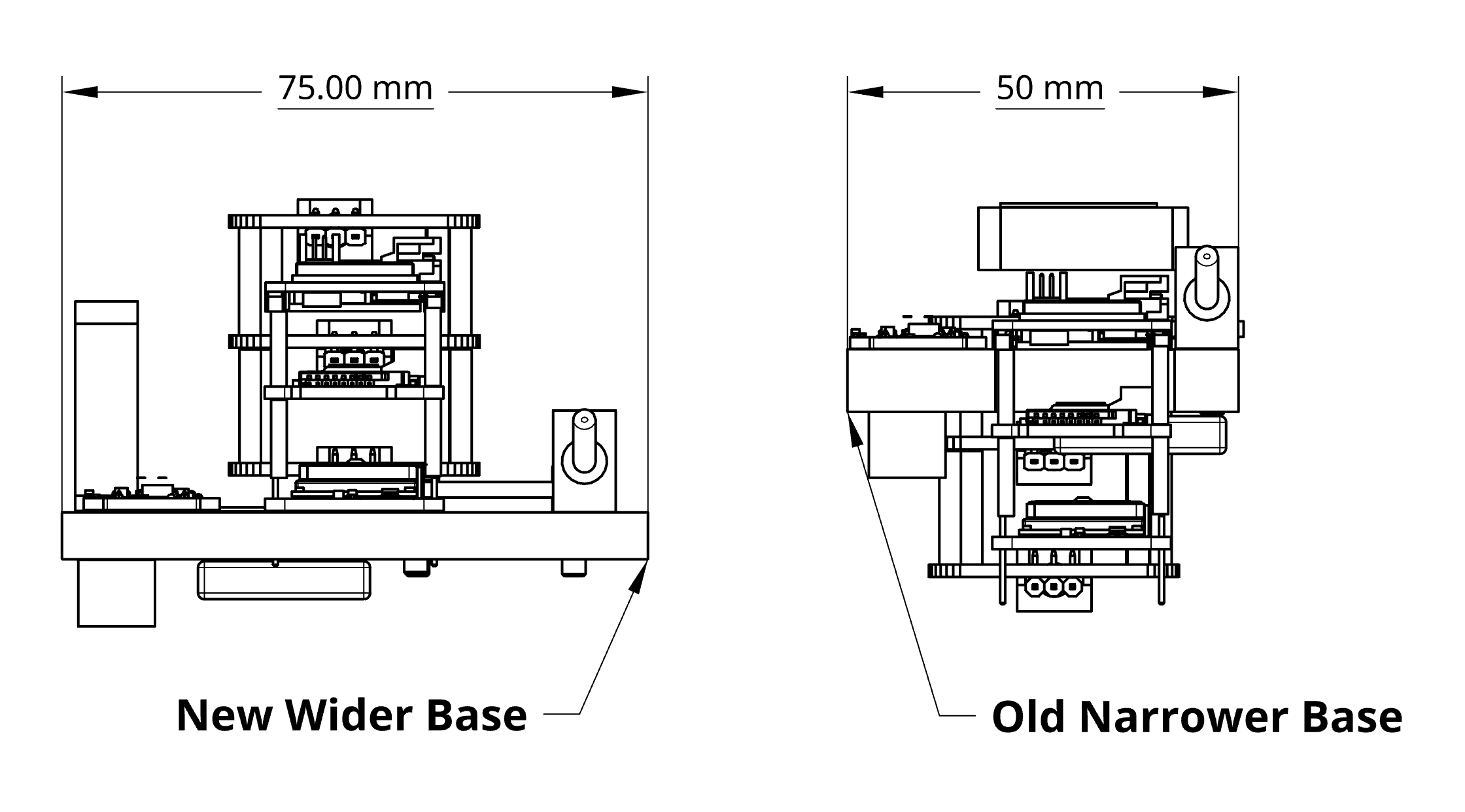}
        \caption{}
        \label{fig:revised-electronics-drawing}
    \end{subfigure}
    \hfill
    \begin{subfigure}[b]{0.30\linewidth}
        \centering
        \includegraphics[width=\linewidth]{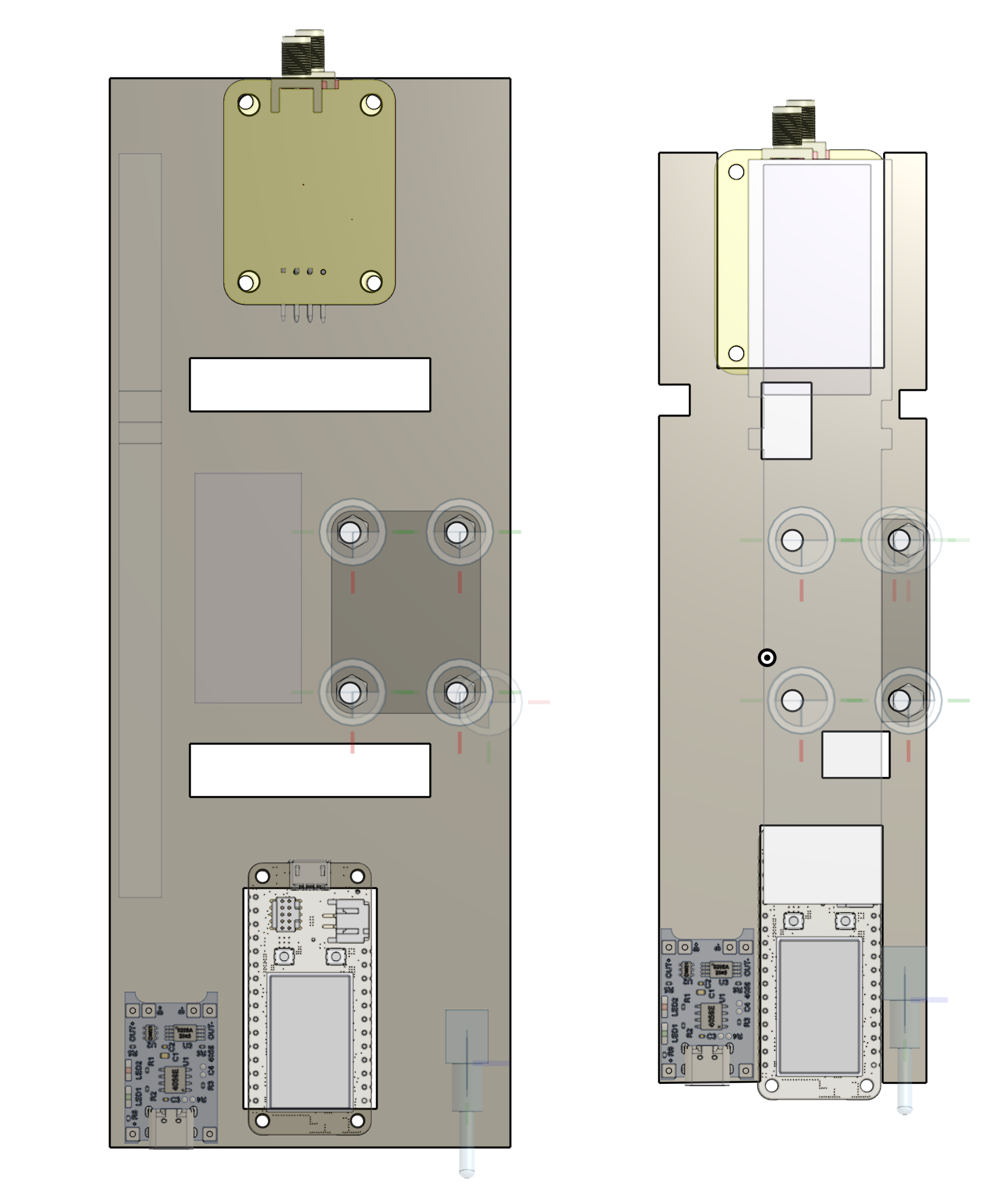}
        \caption{}
        \label{fig:revised-electronics}
    \end{subfigure}
    \caption{(a) shows the wider base and top mounted layout of the newer design. (b) shows the simplified larger MDF board with mounting holes for all of the sensors. This design routes all of the cabling to the underside of the board improving robustness and resilience of the wired connections. }
    \label{fig:row2}
\end{figure*}

Building on the pilot program insights described in Sec.~\ref{sec:Improvements}, we undertook a series of concrete hardware and curriculum modifications targeting the primary failure cases and learning barriers identified during the initial deployment. These changes are designed to improve reproducibility, contain technical complexity for educators and students, and increase overall system robustness.

\subsection{Motivations for Redesign}
\label{subsec:motivation}
\subsubsection{Youth Leader Feedback}
Mentors consistently reported that the most significant friction point during sensor pod assembly was managing the wiring inside the housing. Because the original 2-inch (60\,mm) PVC tube left minimal clearance around the MDF mounting board, inserting the board into the tube generated enough friction to pull Dupont-style connections off sensor pins and the breadboard. Teams that encountered this failure mode in the field were unable to diagnose it without completely disassembling the sensor pod. As kits were distributed in phases, subsequent phases received fully soldered connections to prevent the connection loss issue. While this improved field reliability, it substantially increased build complexity, and that risk to the reliability of the electronics. During the in-person training, youth leaders attended a soldering workshop and the resultant diversity of skill levels across educators motivated the decision to move away from soldering entirely. 

\subsubsection{Component Reliability Analysis}
Post-deployment analysis across the 14 deployed units (Table~\ref{tab:hardware_reliability}) revealed that the turbidity sensor exhibited a 57.1 \% failure rate in the original design, traced to an inadequate seal around the sensor's leak points. A mid-season redesign incorporating an improved seal eliminated all subsequent failures. Other components were mostly reliable given the limited sample size ($n=14$) of teams. 

\subsection{Sensor Pod Redesign}
\label{subsec:pod-redesign}
\subsubsection{Wider Housing}
The primary structural change upgrades the sensor pod
from a 2-inch (60 mm) PVC tube to a 3-inch ($\approx$76 mm) tube, as shown in Fig.~\ref{fig:revised-assembly-drawing}. The enlarged bore provides sufficient clearance so that no sensor cables or Dupont connectors contact the interior walls during board insertion or underwater operation, eliminating the mechanical dislodgment failure mode entirely. With this additional space, all sensor and peripheral connections now use plug-in Dupont connectors. Soldering is now limited exclusively to the power management sub-circuit, which is a contained, well-defined task that can be completed with beginner level soldering experience. 

\subsubsection{Board-Mounted Component Layout}
In the original design, the turbidity sensor board was the only component fastened directly to the MDF mounting board with screws; other components relied on friction-fit or adhesive retention. The redesign extends this board-mount philosophy to all internal components, as illustrated in Fig.~\ref{fig:revised-electronics}. All sensors, microcontrollers, and peripherals are now mechanically fastened to the MDF board with screws before the board is inserted into the tube. This approach ensures consistent structural stability during transport, assembly, and underwater operation, while also keeping the internal layout clearly visible and easy to inspect at every stage of the build.

\subsubsection{Simplified Board Geometry}
The MDF mounting board has been redesigned with a simpler outline that eliminates the complex laser-cut profile required in the original version. The new geometry can be accurately reproduced with a handheld drill alone; a bandsaw, widely available in school carpentry departments, is equally suitable. This change decouples board fabrication from access to specialist equipment, making self-sourcing and local reproduction straightforward for schools and community groups.

\subsection{Training Material Revisions}
\label{subsec:training-revisions}
Concurrent with the hardware redesign, the training and instructional materials have been restructured to adopt a LEGO-style~\cite{Novick2000} visual format. Like LEGO build guides, each stage of construction is presented through clear pictorial steps that are self-contained and produce a visually verifiable intermediate result before the next step begins. This structure reduces cognitive load for both students and educators, directly addressing the engagement gap observed during sensor pod assembly (mean engagement~$= 2.5$) and electronics integration tasks in the pilot. Each task is annotated with an estimated completion time and a difficulty rating, enabling mentors to structure lesson plans more effectively and to set accurate expectations for students with varying technical backgrounds.

\subsection{Lowering the Barrier to Entry}
\label{subsec:barrier}
Taken together, these modifications produce a system that is easier to build, more robust in the field, and more straightforwardly reproducible than its predecessor. Eliminating mandatory soldering from the student assembly workflow removes both a skill prerequisite and a safety concern, opening the program to teams whose mentors have limited electronics experience. The wider housing restores the plug-and-play assembly experience, ensuring that students interact directly with the sensor connections rather than worrying about soldering the connections. The board-mounted layout and simplified MDF geometry mean that a complete sensor pod can be fabricated and assembled using tools available in a typical school workshop, without reliance on expensive laser cutters. The LEGO-style instruction manuals~\cite{LegoLearningThroughPlay} further reinforce this accessibility: by presenting each assembly step as a pictorial action with a clear visual outcome, the materials eliminate the ambiguity that contributed to low engagement during electronics assembly in the pilot. These changes collectively lower the threshold for citizen-science groups and school teams outside the program's direct support network, bringing the Jar~Jar ROV closer to the vision of a community-deployable monitoring platform.   

\bibliographystyle{ieeetr}
\bibliography{bibliography}

\end{document}